\documentclass[sigconf]{acmart}
\AtBeginDocument{%
	\providecommand\BibTeX{{%
			\normalfont B\kern-0.5em{\scshape i\kern-0.25em b}\kern-0.8em\TeX}}}

\usepackage{times}
\usepackage{latexsym}
\usepackage{graphicx}
\usepackage{booktabs}
\usepackage{multirow}
\usepackage{setspace}

\usepackage{mathrsfs}
\usepackage{amsfonts}
\usepackage{amsmath}

\usepackage{algorithm}
\usepackage{algorithmic}
\usepackage[algo2e,ruled,vlined]{algorithm2e}

\copyrightyear{2022} 
\acmYear{2022} 
\setcopyright{acmcopyright}
\acmConference[SIGIR '22]{Proceedings of the 45th International ACM SIGIR Conference on Research and Development in Information Retrieval}{July 11--15, 2022}{Madrid, Spain.}
\acmBooktitle{Proceedings of the 45th International ACM SIGIR Conference on Research and Development in Information Retrieval (SIGIR '22), July 11--15, 2022, Madrid, Spain}
\acmISBN{978-1-4503-8732-3/22/07} 
\acmDOI{10.1145/3477495.3531765}
\begin{document}
\fancyhead{}

\title{Constrained Sequence-to-Tree Generation for Hierarchical Text Classification}

\author{Chao Yu}
\authornote{These authors contributed equally to this work.}
\orcid{0000-0003-1639-5206}
\affiliation{%
	\institution{Alibaba Group}
	\city{Beijing}
	\country{China}}
\email{aiqi.yc@alibaba-inc.com}

\author{Yi Shen}
\authornotemark[1]
\orcid{0000-0001-9706-0867}
\affiliation{%
	\institution{Alibaba Group}
	\city{Beijing}
	\country{China}}
\email{sy133447@alibaba-inc.com}

\author{Yue Mao}
\orcid{0000-0003-0705-7686}
\affiliation{%
	\institution{Alibaba Group}
	\city{Beijing}
	\country{China}}
\email{maoyue.my@alibaba-inc.com}

\author{Longjun Cai}
\affiliation{%
	\institution{Alibaba Group}
	\city{Beijing}
	\country{China}}
\email{longjun.clj@alibaba-inc.com}

\begin{abstract}
	
Hierarchical Text Classification (HTC) is a challenging task where a document can be assigned to multiple hierarchically structured categories within a taxonomy. The majority of prior studies consider HTC as a flat multi-label classification problem, which inevitably leads to ``label inconsistency'' problem. In this paper, we formulate HTC as a sequence generation task and introduce a sequence-to-tree framework (Seq2Tree) for modeling the hierarchical label structure. Moreover, we design a constrained decoding strategy with dynamic vocabulary to secure the label consistency of the results. Compared with previous works, the proposed approach achieves significant and consistent improvements on three benchmark datasets.

\end{abstract}

\ccsdesc[500]{Computing methodologies~Natural language processing}
\ccsdesc[500]{Computing methodologies~Information extraction}
\ccsdesc[500]{Computing methodologies~Natural language generation}

\keywords{text representation, text classification, sequence-to-sequence}


\maketitle

\section{Introduction}

\begin{figure}[]  	
	\centering  
	\includegraphics[scale=0.36]{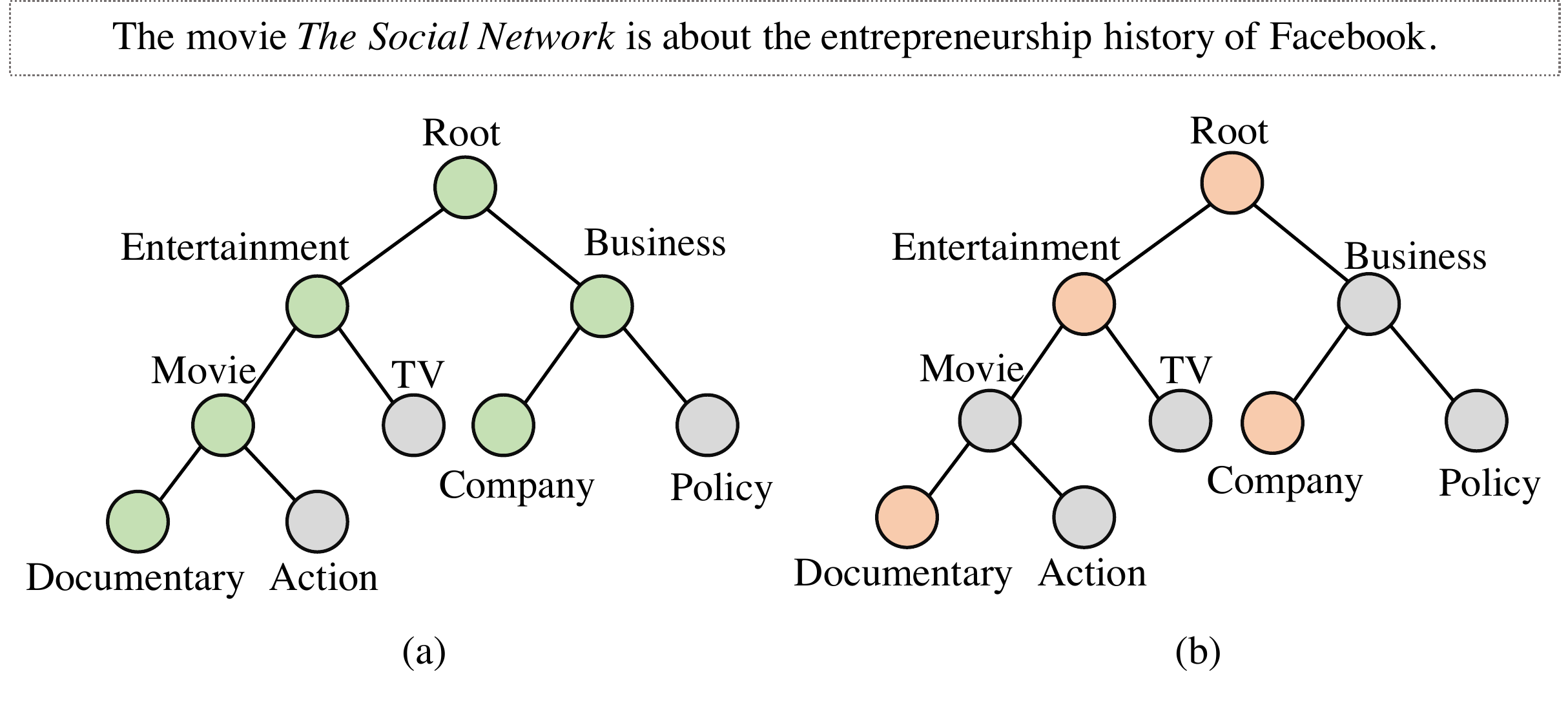}
	\caption{An example of taxonomic hierachy for HTC. (a) The ground truth labels of the input document. (b) An illustration for ``label inconsistency''.  }  
	\label{fig:htc}  	
\end{figure}

Hierarchical text classification (HTC)  is a particular multi-label text classification problem, which aims to assign each document to a set of relevant nodes of a taxonomic hierarchy as depicted in  Figure \ref{fig:htc}(a). HTC has many applications, such as product categorization\cite{cevahir2016large}, fine-grained entity typing \cite{xu2018neural} and news  classification \cite{irsan2019hierarchical}.

Existing approaches can be compartmentalized into two groups: local approaches and global approaches. Local approaches tend to construct multiple local classifiers and usually ignore the holistic structural information of the taxonomic hierarchy. As the mainstream for HTC in recent years, global approaches utilize one single model to deal with all classes and introduce various strategies to capture the hierarchical information of the label space, such as dynamic routing \cite{wang2021concept}, hyperbolic label embedding \cite{chatterjee2021joint}, Tree-LSTM and GCN \cite{zhou2020hierarchy}. However, since all these approaches consider HTC as a flat multi-label classification (MLC) task, the classification results will suffer from ``label inconsistent'' problem.

As depicted in Figure \ref{fig:htc} (a), the ground truth labels of the input document should be the green nodes and these nodes belong to two paths, while traditional MLC approaches may lead to the prediction results as shown in Figure  \ref{fig:htc} (b), where isolated nodes such as \emph{Documentary} and \emph{Company} are produced and they will even be regarded as a valid positive prediction according to traditional evaluation metrics of HTC. In fact, due to the nature of HTC, the prediction of each node should not be in conflict with the results of its ancestors within a path. The isolated predictions and the ``inconsistent paths'' can not meet the needs in many actual application scenarios. In this paper, we aim to address this ``label inconsistent'' problem of HTC.

In order to make the classification results satisfy the ``label consistency'', we should focus on the label dependency within paths in the taxonomic hierarchy instead of each individual node. We associate this with the depth-first traverse algorithm (DFS) \cite{tarjan1972depth} for tree-structure, which is able to ensure that the nodes within the same path could be visited in the top-down order. Inspired by some previous works \cite{vinyals2015grammar, sun2021paradigm,paolini2020structured,wang2018tree} resorting to sequence generation to address various NLP tasks, we propose a sequence to tree (Seq2Tree) framework for HTC. Specifically, we firstly leverage DFS to transform hierarchical labels to a linearized label sequence, then we map the document text and corresponding label sequence in a traditional seq2seq manner. Furthermore, we design a constrained decoding strategy (CD), which can guide the generation process using hierarchical label dependency. In this way, the candidate labels generated at each time step will be limited to the children of the parent node generated at the last time step. In addition to ensuring the label consistency of the results, our approach could make full use of both semantic and structural information of the hierarchical labels during the training process.

In summary, our main contributions are as follows: 
\begin{itemize}
\item We devise a sequence-to-tree generation framework with a constrained decoding strategy for HTC, which effectively guarantees the label consistency of the results.
\item We propose  two new evaluation metrics for HTC task: C-MicroF1 and C-MacroF1, which are more reasonable and more in line with actual application scenarios for HTC. 
\item We conduct experiments on three  benchmark datasets and the results demonstrate that our method outperforms strong baselines on both traditional evaluation metrics and our new proposed metrics. 
\end{itemize}

\section{Methodology}

We aim to leverage sequence generation architecture to address HTC task. To this end, we need to convert the taxonomic hierarchy into a sequence (label linearization). In this section, we start with the problem definition, and then introduce the method for label linearization (\emph{LL}) and the model architecture with constrained decoding strategy (\emph{CD}) in following subsections, respectively.

\subsection{Problem Definition}

We formulate HTC task as $F\text{:}(\mathcal{X},T)\rightarrow \mathcal{Y}$, where $\mathcal{X}\text{=}\{X_{1},...,X_{N}\}$ refers to the set of input documents, $T\text{=}\{V,E\}$ is a predefined  taxonomic hierarchy, $V$ is the set of label nodes and $E$ denotes the parent-child relations between them. $\mathcal{Y}\text{=}\{Y_{1},...,Y_{N}\}$ is the target label sets of $\mathcal{X}$. Our task is to learn a model $F$ that maps a new document $X_{i}\text{=}\{x_{1},...,x_{|X_{i}|}\}$ to its target label set $Y_{i}\text{=}\{y_{1},...,y_{k_{i}}\}$  within $T$, where $k_{i}$ is the number of labels and $k_{i} \leq |V|$.

\subsection{DFS-based Label Linearization}

\begin{figure}[htbp]  	
	\centering  
	\includegraphics[scale=0.3]{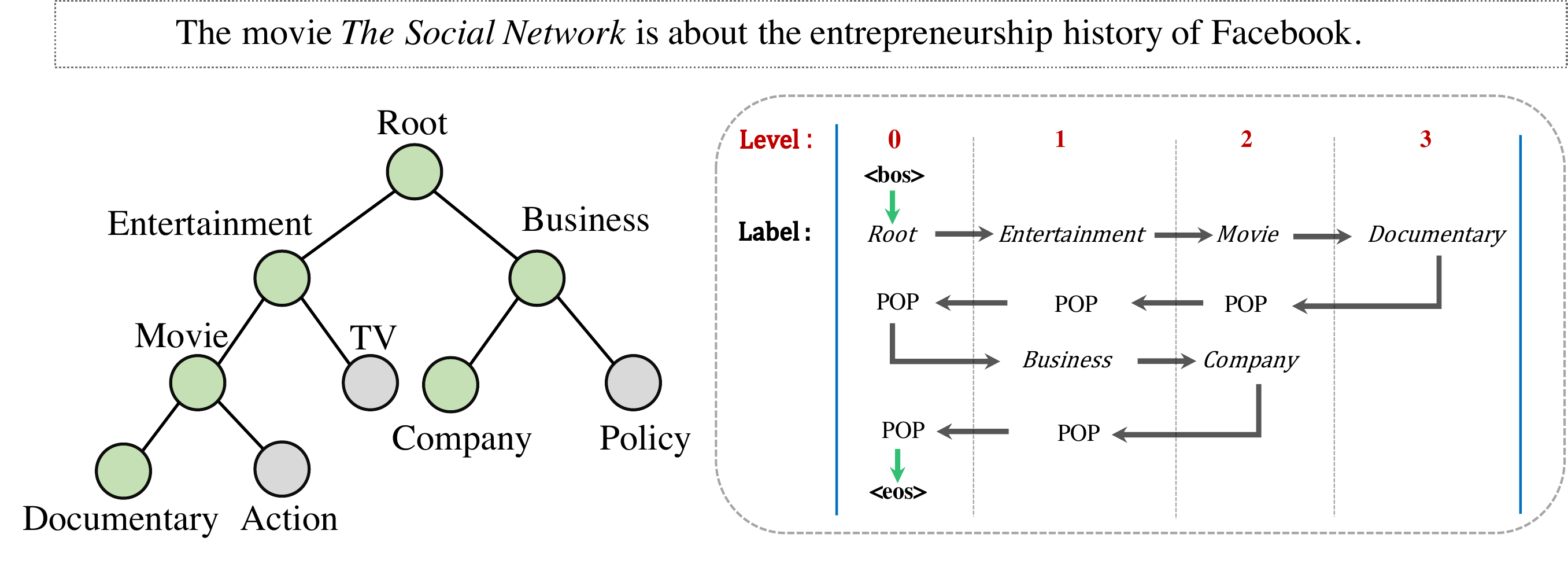}
	\caption{An example of DFS-based label linearization  }  
	\label{fig:dfs}  	
\end{figure}

We achieve label linearization (\emph{LL}) in a very straightforward way following DFS algorithm. As depicted in Figure \ref{fig:dfs}, the target labels $Y_{i}$ of the document make up a subtree of $T$. According to DFS algorithm, these labels could be linearized as $\hat{Y_{i}}$ \{  \emph{Root}-\emph{Entertainment}-\emph{Movie}-\emph{Documentation}-POP-POP-POP-\emph{Business}-\emph{Company}-POP-POP \}, \\where POP is defined as a backtracking operation, it is usually performed right after arriving in the leaf nodes of the subtree. For instance, as in Figure \ref{fig:dfs}, when the leaf node \emph{Documentary} is visited, it is necessary to perform three consecutive POP operations to return to \emph{Root} node for starting a new path \{\emph{Root}-\emph{Business}-\emph{Company}\}. Since DFS algorithm is invertible, DFS-based  \emph{LL} is equivalent to inject the structure information of the hierarchy into the label sequence.

\subsection{Seq2Tree with Constrained Decoding}

\begin{figure*}[htbp]  	
	\centering  
	\includegraphics[scale=0.58]{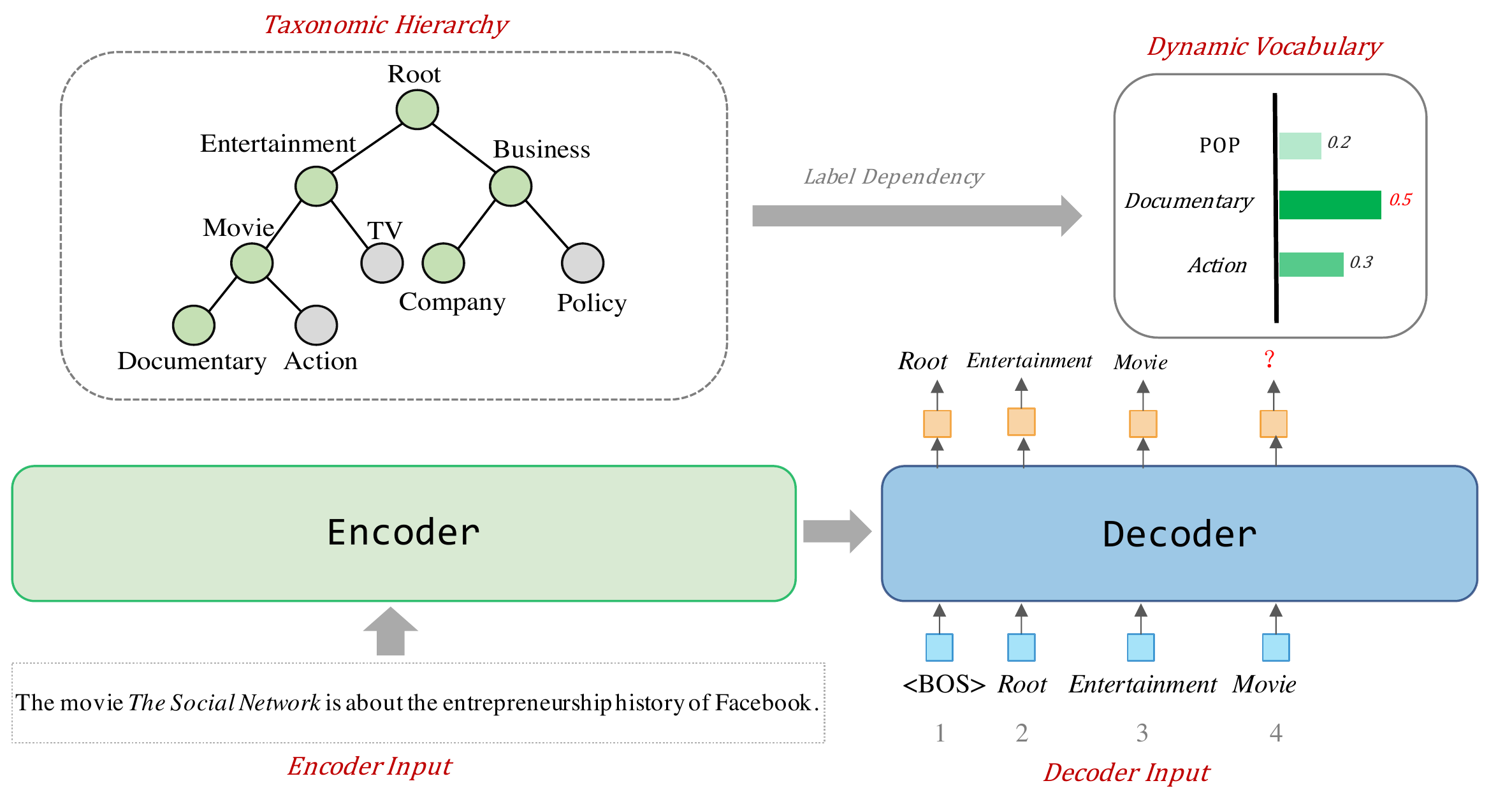}
	\caption{The proposed Seq2Tree framework. }  
	\label{fig:seq2tree}  	
\end{figure*}


As conventional Seq2Seq paradigm, Seq2Tree is also composed of two components: \textbf{Encoder} and \textbf{Decoder}. The architecture is depicted in Figure \ref{fig:seq2tree}.

\textbf{Encoder:} Given the input sequence $X_{i}$, the encoder part is to encode $X_{i}$ into hidden vector representation $H_{i}\text{=}\{h_{1},...,h_{|X_{i}|}\}$ as follows:
\begin{spacing}{0.8}
\begin{equation}
	H_{i}=Encoder(x_{1},...,x_{|X_{i}|})
\end{equation}
\end{spacing}

\textbf{Decoder:}  After the input sequence is encoded, the decoder part predicts the DFS label sequence step by step. Specifically, at time step $i$ of generation, the decoder predicts the $i$-th token $\hat{y_{i}}$ in the linearized DFS label sequence and its corresponding decoder state $h_{i}^{d}$ as:

\begin{equation}
		\hat{y_{i}},h_{i}^{d}= Decoder_{c}(
		[H_{i},h_{<i}^{d}],\hat{y_{i-1}},T) ), \quad \hat{y_{i}} \in DV_{\hat{y_{i}}}
\end{equation}
where $Decoder_{c}$ denotes decoder with \emph{CD}, $h_{<i}^{d}$ denotes decoder state at previous time steps, $\hat{y_{i-1}}$ is the token generated at the last time step. The output sequence starts from the token ``⟨bos⟩'' and ends with the token ``⟨eos⟩''. $DV_{\hat{y_{i}}}$ denotes the dynamic vocabulary for generating $\hat{y_{i}}$, its generation process is elabrated in Algorithm 1: 
\begin{algorithm2e}[]
	\SetAlgoLined
	\KwIn{Generated tokens $\hat{y_{1}},...,\hat{y_{i-1}}$ , Taxonomic hierarchy $T$}
	\KwOut{Dynamic vocabulary $\emph{DV}_{\hat{y_{i}}}$} 
	\BlankLine
	NodeStack $ =$ ()\;
	\For{$j = 0$  \KwTo $i-1$}{
		\eIf{$\hat{y_{j}} \text{ is POP}$}{
			NodeStack.$pop()$\;
		}{
			NodeStack.$add(\hat{y_{j}})$\;
		}
	}
	$\hat{y_{cur}}=$ NodeStack.$pop()$\;
		\eIf{$\hat{y_{cur}} \text{ is Leaf Node}$}{
	$\emph{DV}_{\hat{y_{i}}} = \{POP\}$\;
}{
	$\emph{DV}_{\hat{y_{i}}} = \{child_{T}(\hat{y_{cur}}),POP\}$\;
}
	\BlankLine
	
	\KwRet{$\emph{DV}_{\hat{y_{i}}}$};
	\caption{Dynamic Vocabulary Generation.} \label{algo}
\end{algorithm2e}

When predicting $\hat{y_{i}}$, the \emph{CD} strategy with dynamic vocabulary will be executed. In most cases, the candidate tokens are limited to the child-nodes of $\hat{y_{cur}}$ within $T$ due to the label dependency. As illustrated in Figure \ref{fig:seq2tree}, the model needs to determine which label should be generated at the 4-th decoding time step. Since the model has generated ``\emph{Movie}'' at the last time step, due to the label dependency from $T$, the candidate vocabulary is limited as \{POP, \emph{Documentary}, \emph{Action}\} rather than the whole vocabulary. In this way, we could secure the label consistency during decoding process. 

For the training phase, each instance is a (\emph{text sequence, linearized label sequence}) pair, we define the training object as following negative log likelihood loss:

\begin{spacing}{1.5}
\begin{equation}
	L=-\sum_{(X,\hat{Y})\in \mathcal{(X,\hat{Y})}} log(P_{\theta}(\hat{Y}|X,T))
\end{equation}
\begin{equation}
	P_{\theta}(\hat{Y}|X,T) =  \prod_{i}^{|\hat{Y}|}P_{\theta}(\hat{y_{i}}|\hat{y_{<i}},X,T)
\end{equation}
\end{spacing}

where $\theta$ is model parameters, $\hat{y_{<i}}$ denotes the previous generated labels. $P_{\theta}(\hat{y_{i}}|\hat{y_{<i}},X,T)$ is the probability distribution over the dynamic candidate vocabulary $\emph{DV}_{\hat{y_{i}}}$, which is normalized by $softmax(\cdot)$. 

For the inference phase, we resort to beam search strategy to generate the output label sequence as:

\begin{spacing}{1}

		\begin{equation}
			\begin{split}
				\hat{Y^{*}}&=\mathop{\arg\max}_{\hat{Y}}log(P_{\theta^{*}}(\hat{Y}|X,T))\\
				&=\mathop{\arg\max}_{\hat{Y}:\{\hat{y_{1}},...\hat{y_{k_{i}}}\}}log\prod_{i}^{|\hat{Y}|}P_{\theta^{*}}(\hat{y_{i}}|\hat{y_{<i}},X,T)
			\end{split}
		\end{equation} \label{eq:inference}

\end{spacing}

 Finally, we obtain $\hat{Y^*}=\{\hat{y_{1}}^{*},...,\hat{y_{k_{i}}}^{*}\}$ as the predicted label sequence. We adopt T5-Base \cite{raffel2020exploring} as the backbone to implement Seq2Tree. More implementation details are reported in \textbf{Section 3.1}.

\section{Experiment}

\begin{table*}[]
	\footnotesize
		\caption{ Results of different methods on three datasets. The results of SGM and HiLAP-RL are reported by \citet{zhou2020hierarchy}. The results of other baselines are from corresponding original papers. ``-'' means not available in the original paper. Our implementation results are marked by ``*''. ``HiMatch-BERT+pp'' is an abbreviation of ``HiMatch-BERT+post processing''.  Since HiLAP-RL, HiMatch-BERT+pp and our method have guaranteed the label consistency, the results of them on each metric and its corresponding path constrained variant are always the same. ``${\uparrow}$'' indicates the improvement of our method compared with HiMatch-BERT.}
	\setlength{\tabcolsep}{1mm}{
		
		\begin{tabular}{@{}cllllllllllll@{}}
			\toprule
			\multirow{2}{*}{Model} & \multicolumn{4}{c}{\textbf{RCV1-V2}}                                              & \multicolumn{4}{c}{\textbf{WOS}}                                                  & \multicolumn{4}{c}{\textbf{BGC}}                                  \\ \cmidrule(l){2-13} 
			& Micro F1       & C-Micro F1      & Macro F1       & \multicolumn{1}{l|}{C-Macro F1} & Micro F1       & C-Micro F1     & Macro F1       & \multicolumn{1}{l|}{C-Macro F1} & Micro F1       & C-Micro F1      & Macro F1       & C-Macro F1     \\ \midrule
			HMC-Capsule                  & -         & -       & -          & \multicolumn{1}{l|}{ -   }                      & -         & -        & -          & \multicolumn{1}{l|}{ - }                        & 74.37         & -         & -         & -        \\
			SGM                  & 77.30         & -         & 47.49         & \multicolumn{1}{l|}{ -  }                       & -       & -        & -         &\multicolumn{1}{l|}{  -  }                      &-         & -      &-        & -         \\
			HiLAP-RL                  & 83.30         & 83.30         & 60.10         &\multicolumn{1}{l|}{ 60.10   }                       & -       & -        & -         &\multicolumn{1}{l|}{  -  }                      &-         & -      &-        & -         \\
			HTCInfoMax                  & 83.51       & -         & 62.71        & \multicolumn{1}{l|}{ -  }                       & 85.58       & -        & 80.05         &\multicolumn{1}{l|}{  -  }                      &-         & -      &-        & -         \\
			SGM-T5                  & 84.39*        & 83.75*         & 65.09*        & \multicolumn{1}{l|}{ 64.55*}                       & 85.83*       & 85.14*        & 80.79*         &\multicolumn{1}{l|}{  79.95*  }                      &77.84*         & 76.72*      &60.91*        & 59.97*         \\
			HiAGM                  & 83.96          & 83.05          & 63.35          &\multicolumn{1}{l|}{59.64}                       & 85.82          & 85.35          & 80.28          & \multicolumn{1}{l|}{ 79.84  }                        & 77.22*          & 76.16*          & 57.91*          & 56.61*          \\
			HiMatch                & 84.73          & 83.49          & 64.11          & \multicolumn{1}{l|}{60.64}                         & 86.20           & 85.61          & 80.53          &\multicolumn{1}{l|}{  79.32  }                        & 76.57*          & 75.23*          & 58.34*          & 56.31*          \\
			HiMatch-BERT           & 86.33          & 85.25          & 68.66          &\multicolumn{1}{l|}{ 67.15}                         & 86.70          & 85.74          & 81.06          &\multicolumn{1}{l|}{  79.86   }                       & 78.89*          & 78.01*          & 63.19*          & 62.23*          \\
			
			HiMatch-BERT +  pp         & 85.37*          & 85.37*          & 67.01*          &\multicolumn{1}{l|}{ 67.01*}                         & 85.86*          & 85.86*         & 80.18*          &\multicolumn{1}{l|}{  80.18*    }                       & 77.59*          & 77.59*          & 62.05*          & 62.05*          \\
			Seq2Tree(Ours)                  & 
			\textbf{86.88}$^{\uparrow0.55}$& \textbf{86.88}$^{\uparrow1.63}$ & \textbf{70.01}$^{\uparrow1.35}$ & \multicolumn{1}{l|}{ \textbf{70.01}$^{\uparrow2.86}$} & \textbf{87.20}$^{\uparrow0.50}$ & \textbf{87.20}$^{\uparrow1.46}$ & \textbf{82.50}$^{\uparrow1.44}$ &\multicolumn{1}{l|}{  \textbf{82.50}$^{\uparrow2.64}$ }& \textbf{79.72}$^{\uparrow0.83}$ & \textbf{79.72}$^{\uparrow1.71}$ & \textbf{63.96}$^{\uparrow0.77}$ & \textbf{63.96}$^{\uparrow1.73}$ \\ \bottomrule
	\end{tabular}}

	\label{tab:main_result}
\end{table*}

\subsection{Experimental Settings}

\textbf{Datasets}: We evaluate our method on three widely used datasets for HTC, including RCV1-V2\footnote{\url{http://www.ai.mit.edu/projects/jmlr/papers/volume5/lewis04a/lyrl2004\_rcv1v2\_README.htm}}, WOS\footnote{\url{https://data.mendeley.com/datasets/9rw3vkcfy4/2}} and BGC\footnote{\url{https://www.inf.uni-hamburg.de/en/inst/ab/lt/resources/data/blurb-genre-collection.html}}. 
The categories of these datasets are all structured as a tree-like hierarchy. We keep exactly the same data splits as reported in the previous works \cite{zhou2020hierarchy,chen2021hierarchy}. It is worth noting that WOS is for single-path HTC while RCV1-V2 and BGC are both for multi-path HTC.
Please refer to Table \ref{tab:datasets_statics} for the detailed statistics of the datasets.


\textbf{Baseline}: We compare our method with following HTC models: SGM \cite{yang2018sgm}, HMC-capsule \cite{aly2019hierarchical},  HiLAP-RL \cite{mao2019hierarchical}, HiAGM \cite{zhou2020hierarchy}, HTCInfoMax \cite{deng2021htcinfomax}, HiMatch and HiMatch-BERT \cite{chen2021hierarchy}\footnote{Please refer to Section 4 for introduction of these baseline methods.}. Since SGM is also a generative method for HTC, we implement a variant (SGM-T5) of SGM for a fair comparison, which replaced its LSTM-based encoder-decoder backbone with T5. Among these baselines, HiMatch-BERT is the state-of-the-art model. Besides, HiLAP-RL  \cite{mao2019hierarchical} is the only method that attempts to address the ``label inconsistency'' problem, which leverages deep reinforcement learning to tackle this issue. In \cite{mao2019hierarchical}, the authors also mention that ``label inconsistency'' problem could be solved through post processing operation, i.e., simply assign the ancestors of the isolated nodes to the prediction results of corresponding isolated nodes. Therefore, we also implement this post processing on HiMatch-BERT to validate its effectiveness.

\textbf{Evaluation Metrics}: Besides Micro-F1 and Macro-F1, which are widely adopted evaluation metrics in existing HTC studies \cite{aly2019hierarchical, zhou2020hierarchy, deng2021htcinfomax, chen2021hierarchy}, we propose two new metrics: path-constrained MicroF1 (C-MicroF1) and path-constrained MacroF1 (C-MacroF1). The difference between these path-constrained variants and traditional metrics is that, the prediction result for a node will be regarded  as ``true'' only when all its ancestor nodes have been predicted as ``true''. Take Figure \ref{fig:htc}(b) as an example, although the model has made correct predictions for \emph{Documentary} and \emph{Company}, the output results for these two nodes are still regarded as ``false'' due to mistakes happened on their parent nodes.

\textbf{Implementation Details}: For all datasets, the batch size is set to 16 and we optimize the model with label smoothing \cite{muller2019does} and AdamW \cite{loshchilov2018decoupled} with a learning rate of 5e-5. The results reported in our experiments are all based on the average score of 5 runs with different random seeds. For baseline models (HiAGM, HiMatch and HiMatch-Bert), we use the implementations provided by the authors to produce the experimental results on BGC dataset, the hyperparameters of these baseline models are manually tuned on the standard validation set provided in BGC. Our experiments are all conducted on a single Tesla V100M32 GPU. Besides, during the decoding stage, the tokens already generated previously are also excluded in subsequential dynamic vocabularies.

\subsection{Main Results}

The experimental results are elaborated in Table \ref{tab:main_result} and our proposed Seq2Tree consistently outperforms previous approaches across all the datasets. Compared with SGM-T5, Seq2Tree also outperforms it with a large margin, which validates that the improvement of our method is mainly brought by our design on the framework rather than T5. In general, the performance improvements on C-MicroF1 and C-MacroF1 are more remarkable than on MicroF1 and MacroF1. The reason is that Seq2Tree is able to guarantee the label consistency on each path through the \emph{CD} strategy. It is also notable that the performance of post-processing  does not improved but decreased. The reason is that HiMatch-BERT cannot guarantee the prediction accuracy of the isolated nodes, and simply assigning ancestor nodes for documents has introduced a lot of noise, which causes the performance degradation of HiMatch-BERT+pp.


\begin{table}[]
	\vspace{-0.5em}
	\centering
	\small
	\caption{ The statistics of datasets. \emph{|V|} is the total number of labels.  \emph{Depth} is the maximum level of the label hierarchy. \emph{ Avg(|$V$|)} is the average number of labels in each sample.
	}
	\begin{tabular}{@{}ccccccc@{}}
		\toprule
		Dataset & $|V|$ & Depth & Avg(|$V$|) & Train & Validation   & Test   \\ 
		\midrule
		RCV1-V2 & 103 & 4     & 3.24     & 20833 & 2316  & 781265 \\
		WOS     & 141 & 2     & 2        & 30070 & 7518  & 9397   \\
		BGC     & 146 & 4     & 3.01     & 58715 & 14785 & 18394  \\ 
		\bottomrule
	\end{tabular}
	
	\label{tab:datasets_statics}
\end{table}

\subsection{Ablation Study}

To reveal the individual effects of label linearization (\emph{LL}) and constrained decoding strategy (\emph{CD}), we implement different variants of Seq2Tree by removing \emph{CD} and \emph{LL} sequentially. The experimental results on RCV1-V2 are reported in Table \ref{tab:ablation_study}. Removing \emph{CD} results in  significant performance degradation, which validates its effectiveness. The performance will further decrease when we remove both  \emph{CD} and \emph{LL}, which illustrating the importance of incorporating the holistic tree-structural information of the taxonomic hierarchy through DFS.
\begin{table}[ht]
	\vspace{-0.5em}
	\small
	\centering
		\caption{Ablation study results on RCV1-V2. When both \emph{CD} and \emph{LL} are removed, Seq2Tree is equivalent to SGM-T5. }
	\begin{tabular}{@{}ccccccc@{}}
		\toprule
		Model & Micro F1       & C-Micro F1      & Macro F1       & C-Macro F1      \\ \midrule
		Seq2Tree       & \textbf{86.88} & \textbf{86.88} & \textbf{70.01} & \textbf{70.01} \\
		w/o CD         & 85.31          & 84.98          & 66.23          & 65.15          \\
		w/o CD\&LL   & 84.39          & 83.75          & 65.09          & 64.55          \\ \bottomrule
	\end{tabular}

	\label{tab:ablation_study}
\end{table}

\section{Related Work}

Existing works for HTC could be categorized into local and global approaches. Local approaches \cite{cesa2006hierarchical,shimura2018hft,banerjee2019hierarchical} tend to construct multiple local classifiers and usually ignore the holistic structural information of the taxonomic hierarchy. As the mainstream for HTC in recent years, global approaches utilize one single model to deal with all classes and introduce various strategies to capture the hierarchical information of the label space. For instance, \cite{wang2021concept} adopt a hierarchical network to extract the concepts and model the sharing process via a modified dynamic routing algorithm. HMC-Capsule \cite{aly2019hierarchical} employs capsule networks to classify documents into hierarchical structured labels. HiLAP-RL \cite{mao2019hierarchical} formulates HTC as a Markov decision process and propose to learn a label assignment policy via deep reinforcement learning. HTCInfoMax \cite{deng2021htcinfomax} utilizes text-label mutual information maximization algorithm and label prior matching strategy for capturing hierarchical information between labels. In HiAGM \cite{zhou2020hierarchy}, two structure encoders (Tree-LSTM and GCN) are introduced for modeling hierarchical label in both top-down and bottom-up manners. HiMatch \cite{chen2021hierarchy} proposes a hierarchy-aware label semantics matching network to learn the text-label semantics matching relationship in a hierarchy-aware manner. HVHMC \cite{xu2021hierarchical} introduces loosely coupled graph convolutional neural network as the representation component for capturing vertical and horizontal label dependencies simultaneously. Although the methods mentioned above have been successful to a certain extent, most of them suffer from the ``label inconsistency'' problem. 

It is worth mentioning that there are two previous works also resort to sequence-to-sequence paradigm to address HTC task, one is SGM \cite{yang2018sgm}, the other is  \cite{risch2020hierarchical}.  However, these two works are limited to single path scenarios which can not tackle the multi-path cases. Moreover, since these two methods generate hierarchical labels sequence uncontrollably, which will also suffer from ``label inconsistent'' problem.

\section{Conclusion}

In this paper, we propose Seq2Tree method for addressing the ``label inconsistency'' limitation of prior HTC approaches. Seq2Tree formulates HTC as a sequence generation problem and adopts a T5-based seq2seq architecture to generate a DFS-styled label sequence for each document. In this simple but effective manner, Seq2Tree is able to capture the structural information of the taxonomic hierachy. It secures the consistency of generated labels through the constrained decoding strategy with dynamic vocabulary. The experimental results on three benchmark datasets demonstrate the effectiveness of Seq2Tree.


\bibliographystyle{ACM-Reference-Format}
\bibliography{main}


\begin{thebibliography}{24}


\ifx \showCODEN    \undefined \def \showCODEN     #1{\unskip}     \fi
\ifx \showDOI      \undefined \def \showDOI       #1{#1}\fi
\ifx \showISBNx    \undefined \def \showISBNx     #1{\unskip}     \fi
\ifx \showISBNxiii \undefined \def \showISBNxiii  #1{\unskip}     \fi
\ifx \showISSN     \undefined \def \showISSN      #1{\unskip}     \fi
\ifx \showLCCN     \undefined \def \showLCCN      #1{\unskip}     \fi
\ifx \shownote     \undefined \def \shownote      #1{#1}          \fi
\ifx \showarticletitle \undefined \def \showarticletitle #1{#1}   \fi
\ifx \showURL      \undefined \def \showURL       {\relax}        \fi
\providecommand\bibfield[2]{#2}
\providecommand\bibinfo[2]{#2}
\providecommand\natexlab[1]{#1}
\providecommand\showeprint[2][]{arXiv:#2}

\bibitem[\protect\citeauthoryear{Aly, Remus, and Biemann}{Aly
  et~al\mbox{.}}{2019}]%
        {aly2019hierarchical}
\bibfield{author}{\bibinfo{person}{Rami Aly}, \bibinfo{person}{Steffen Remus},
  {and} \bibinfo{person}{Chris Biemann}.} \bibinfo{year}{2019}\natexlab{}.
\newblock \showarticletitle{Hierarchical multi-label classification of text
  with capsule networks}. In \bibinfo{booktitle}{\emph{Proceedings of the 57th
  Annual Meeting of the Association for Computational Linguistics: Student
  Research Workshop}}. \bibinfo{pages}{323--330}.
\newblock


\bibitem[\protect\citeauthoryear{Banerjee, Akkaya, Perez-Sorrosal, and
  Tsioutsiouliklis}{Banerjee et~al\mbox{.}}{2019}]%
        {banerjee2019hierarchical}
\bibfield{author}{\bibinfo{person}{Siddhartha Banerjee}, \bibinfo{person}{Cem
  Akkaya}, \bibinfo{person}{Francisco Perez-Sorrosal}, {and}
  \bibinfo{person}{Kostas Tsioutsiouliklis}.} \bibinfo{year}{2019}\natexlab{}.
\newblock \showarticletitle{Hierarchical transfer learning for multi-label text
  classification}. In \bibinfo{booktitle}{\emph{Proceedings of the 57th Annual
  Meeting of the Association for Computational Linguistics}}.
  \bibinfo{pages}{6295--6300}.
\newblock


\bibitem[\protect\citeauthoryear{Cesa-Bianchi, Gentile, and
  Zaniboni}{Cesa-Bianchi et~al\mbox{.}}{2006}]%
        {cesa2006hierarchical}
\bibfield{author}{\bibinfo{person}{Nicolo Cesa-Bianchi},
  \bibinfo{person}{Claudio Gentile}, {and} \bibinfo{person}{Luca Zaniboni}.}
  \bibinfo{year}{2006}\natexlab{}.
\newblock \showarticletitle{Hierarchical classification: combining bayes with
  svm}. In \bibinfo{booktitle}{\emph{Proceedings of the 23rd international
  conference on Machine learning}}. \bibinfo{pages}{177--184}.
\newblock


\bibitem[\protect\citeauthoryear{Cevahir and Murakami}{Cevahir and
  Murakami}{2016}]%
        {cevahir2016large}
\bibfield{author}{\bibinfo{person}{Ali Cevahir} {and} \bibinfo{person}{Koji
  Murakami}.} \bibinfo{year}{2016}\natexlab{}.
\newblock \showarticletitle{Large-scale Multi-class and Hierarchical Product
  Categorization for an E-commerce Giant}. In
  \bibinfo{booktitle}{\emph{Proceedings of the 26th International Conference on
  Computational Linguistics: Technical Papers}}. \bibinfo{pages}{525--535}.
\newblock


\bibitem[\protect\citeauthoryear{Chatterjee, Maheshwari, Ramakrishnan, and
  Jagaralpudi}{Chatterjee et~al\mbox{.}}{2021}]%
        {chatterjee2021joint}
\bibfield{author}{\bibinfo{person}{Soumya Chatterjee}, \bibinfo{person}{Ayush
  Maheshwari}, \bibinfo{person}{Ganesh Ramakrishnan}, {and}
  \bibinfo{person}{Saketha~Nath Jagaralpudi}.} \bibinfo{year}{2021}\natexlab{}.
\newblock \showarticletitle{Joint Learning of Hyperbolic Label Embeddings for
  Hierarchical Multi-label Classification}. In
  \bibinfo{booktitle}{\emph{Proceedings of the 16th Conference of the European
  Chapter of the Association for Computational Linguistics}}.
  \bibinfo{pages}{2829--2841}.
\newblock


\bibitem[\protect\citeauthoryear{Chen, Ma, Lin, and Yan}{Chen
  et~al\mbox{.}}{2021}]%
        {chen2021hierarchy}
\bibfield{author}{\bibinfo{person}{Haibin Chen}, \bibinfo{person}{Qianli Ma},
  \bibinfo{person}{Zhenxi Lin}, {and} \bibinfo{person}{Jiangyue Yan}.}
  \bibinfo{year}{2021}\natexlab{}.
\newblock \showarticletitle{Hierarchy-aware Label Semantics Matching Network
  for Hierarchical Text Classification}. In
  \bibinfo{booktitle}{\emph{Proceedings of the 59th Annual Meeting of the
  Association for Computational Linguistics and the 11th International Joint
  Conference on Natural Language Processing}}. \bibinfo{pages}{4370--4379}.
\newblock


\bibitem[\protect\citeauthoryear{Deng, Peng, He, Li, and Philip}{Deng
  et~al\mbox{.}}{2021}]%
        {deng2021htcinfomax}
\bibfield{author}{\bibinfo{person}{Zhongfen Deng}, \bibinfo{person}{Hao Peng},
  \bibinfo{person}{Dongxiao He}, \bibinfo{person}{Jianxin Li}, {and}
  \bibinfo{person}{S~Yu Philip}.} \bibinfo{year}{2021}\natexlab{}.
\newblock \showarticletitle{HTCInfoMax: A Global Model for Hierarchical Text
  Classification via Information Maximization}. In
  \bibinfo{booktitle}{\emph{Proceedings of the 2021 Conference of the North
  American Chapter of the Association for Computational Linguistics: Human
  Language Technologies}}. \bibinfo{pages}{3259--3265}.
\newblock


\bibitem[\protect\citeauthoryear{Irsan and Khodra}{Irsan and Khodra}{2019}]%
        {irsan2019hierarchical}
\bibfield{author}{\bibinfo{person}{Ivana~Clairine Irsan} {and}
  \bibinfo{person}{Masayu~Leylia Khodra}.} \bibinfo{year}{2019}\natexlab{}.
\newblock \showarticletitle{Hierarchical multi-label news article
  classification with distributed semantic model based features}.
\newblock \bibinfo{journal}{\emph{International Journal of Advances in
  Intelligent Informatics}} \bibinfo{volume}{5}, \bibinfo{number}{1}
  (\bibinfo{year}{2019}), \bibinfo{pages}{40--47}.
\newblock


\bibitem[\protect\citeauthoryear{Loshchilov and Hutter}{Loshchilov and
  Hutter}{2018}]%
        {loshchilov2018decoupled}
\bibfield{author}{\bibinfo{person}{Ilya Loshchilov} {and}
  \bibinfo{person}{Frank Hutter}.} \bibinfo{year}{2018}\natexlab{}.
\newblock \showarticletitle{Decoupled Weight Decay Regularization}. In
  \bibinfo{booktitle}{\emph{International Conference on Learning
  Representations}}.
\newblock


\bibitem[\protect\citeauthoryear{Mao, Tian, Han, and Ren}{Mao
  et~al\mbox{.}}{2019}]%
        {mao2019hierarchical}
\bibfield{author}{\bibinfo{person}{Yuning Mao}, \bibinfo{person}{Jingjing
  Tian}, \bibinfo{person}{Jiawei Han}, {and} \bibinfo{person}{Xiang Ren}.}
  \bibinfo{year}{2019}\natexlab{}.
\newblock \showarticletitle{Hierarchical Text Classification with Reinforced
  Label Assignment}. In \bibinfo{booktitle}{\emph{Proceedings of the 2019
  Conference on Empirical Methods in Natural Language Processing and the 9th
  International Joint Conference on Natural Language Processing}}.
  \bibinfo{pages}{445--455}.
\newblock


\bibitem[\protect\citeauthoryear{M{\"u}ller, Kornblith, and Hinton}{M{\"u}ller
  et~al\mbox{.}}{2019}]%
        {muller2019does}
\bibfield{author}{\bibinfo{person}{Rafael M{\"u}ller}, \bibinfo{person}{Simon
  Kornblith}, {and} \bibinfo{person}{Geoffrey Hinton}.}
  \bibinfo{year}{2019}\natexlab{}.
\newblock \showarticletitle{When Does Label Smoothing Help?}
\newblock \bibinfo{journal}{\emph{arXiv preprint arXiv:1906.02629}}
  (\bibinfo{year}{2019}).
\newblock


\bibitem[\protect\citeauthoryear{Paolini, Athiwaratkun, Krone, Ma, Achille,
  ANUBHAI, dos Santos, Xiang, and Soatto}{Paolini et~al\mbox{.}}{2020}]%
        {paolini2020structured}
\bibfield{author}{\bibinfo{person}{Giovanni Paolini}, \bibinfo{person}{Ben
  Athiwaratkun}, \bibinfo{person}{Jason Krone}, \bibinfo{person}{Jie Ma},
  \bibinfo{person}{Alessandro Achille}, \bibinfo{person}{RISHITA ANUBHAI},
  \bibinfo{person}{Cicero~Nogueira dos Santos}, \bibinfo{person}{Bing Xiang},
  {and} \bibinfo{person}{Stefano Soatto}.} \bibinfo{year}{2020}\natexlab{}.
\newblock \showarticletitle{Structured Prediction as Translation between
  Augmented Natural Languages}. In \bibinfo{booktitle}{\emph{International
  Conference on Learning Representations}}.
\newblock


\bibitem[\protect\citeauthoryear{Raffel, Shazeer, Roberts, Lee, Narang, Matena,
  Zhou, Li, and Liu}{Raffel et~al\mbox{.}}{2020}]%
        {raffel2020exploring}
\bibfield{author}{\bibinfo{person}{Colin Raffel}, \bibinfo{person}{Noam
  Shazeer}, \bibinfo{person}{Adam Roberts}, \bibinfo{person}{Katherine Lee},
  \bibinfo{person}{Sharan Narang}, \bibinfo{person}{Michael Matena},
  \bibinfo{person}{Yanqi Zhou}, \bibinfo{person}{Wei Li}, {and}
  \bibinfo{person}{Peter~J Liu}.} \bibinfo{year}{2020}\natexlab{}.
\newblock \showarticletitle{Exploring the Limits of Transfer Learning with a
  Unified Text-to-Text Transformer}.
\newblock \bibinfo{journal}{\emph{Journal of Machine Learning Research}}
  \bibinfo{volume}{21} (\bibinfo{year}{2020}), \bibinfo{pages}{1--67}.
\newblock


\bibitem[\protect\citeauthoryear{Risch, Garda, and Krestel}{Risch
  et~al\mbox{.}}{2020}]%
        {risch2020hierarchical}
\bibfield{author}{\bibinfo{person}{Julian Risch}, \bibinfo{person}{Samuele
  Garda}, {and} \bibinfo{person}{Ralf Krestel}.}
  \bibinfo{year}{2020}\natexlab{}.
\newblock \showarticletitle{Hierarchical document classification as a sequence
  generation task}. In \bibinfo{booktitle}{\emph{Proceedings of the ACM/IEEE
  Joint Conference on Digital Libraries in 2020}}. \bibinfo{pages}{147--155}.
\newblock


\bibitem[\protect\citeauthoryear{Shimura, Li, and Fukumoto}{Shimura
  et~al\mbox{.}}{2018}]%
        {shimura2018hft}
\bibfield{author}{\bibinfo{person}{Kazuya Shimura}, \bibinfo{person}{Jiyi Li},
  {and} \bibinfo{person}{Fumiyo Fukumoto}.} \bibinfo{year}{2018}\natexlab{}.
\newblock \showarticletitle{HFT-CNN: Learning hierarchical category structure
  for multi-label short text categorization}. In
  \bibinfo{booktitle}{\emph{Proceedings of the 2018 Conference on Empirical
  Methods in Natural Language Processing}}. \bibinfo{pages}{811--816}.
\newblock


\bibitem[\protect\citeauthoryear{Sun, Liu, Qiu, and Huang}{Sun
  et~al\mbox{.}}{2021}]%
        {sun2021paradigm}
\bibfield{author}{\bibinfo{person}{Tianxiang Sun}, \bibinfo{person}{Xiangyang
  Liu}, \bibinfo{person}{Xipeng Qiu}, {and} \bibinfo{person}{Xuanjing Huang}.}
  \bibinfo{year}{2021}\natexlab{}.
\newblock \showarticletitle{Paradigm Shift in Natural Language Processing}.
\newblock \bibinfo{journal}{\emph{arXiv preprint arXiv:2109.12575}}
  (\bibinfo{year}{2021}).
\newblock


\bibitem[\protect\citeauthoryear{Tarjan}{Tarjan}{1972}]%
        {tarjan1972depth}
\bibfield{author}{\bibinfo{person}{Robert Tarjan}.}
  \bibinfo{year}{1972}\natexlab{}.
\newblock \showarticletitle{Depth-first search and linear graph algorithms}.
\newblock \bibinfo{journal}{\emph{SIAM journal on computing}}
  \bibinfo{volume}{1}, \bibinfo{number}{2} (\bibinfo{year}{1972}),
  \bibinfo{pages}{146--160}.
\newblock


\bibitem[\protect\citeauthoryear{Vinyals, Kaiser, Koo, Petrov, Sutskever, and
  Hinton}{Vinyals et~al\mbox{.}}{2015}]%
        {vinyals2015grammar}
\bibfield{author}{\bibinfo{person}{Oriol Vinyals}, \bibinfo{person}{{\L}ukasz
  Kaiser}, \bibinfo{person}{Terry Koo}, \bibinfo{person}{Slav Petrov},
  \bibinfo{person}{Ilya Sutskever}, {and} \bibinfo{person}{Geoffrey Hinton}.}
  \bibinfo{year}{2015}\natexlab{}.
\newblock \showarticletitle{Grammar as a foreign language}.
\newblock \bibinfo{journal}{\emph{Advances in neural information processing
  systems}}  \bibinfo{volume}{28} (\bibinfo{year}{2015}),
  \bibinfo{pages}{2773--2781}.
\newblock


\bibitem[\protect\citeauthoryear{Wang, Pham, Yin, and Neubig}{Wang
  et~al\mbox{.}}{2018}]%
        {wang2018tree}
\bibfield{author}{\bibinfo{person}{Xinyi Wang}, \bibinfo{person}{Hieu Pham},
  \bibinfo{person}{Pengcheng Yin}, {and} \bibinfo{person}{Graham Neubig}.}
  \bibinfo{year}{2018}\natexlab{}.
\newblock \showarticletitle{A Tree-based Decoder for Neural Machine
  Translation}. In \bibinfo{booktitle}{\emph{Proceedings of the 2018 Conference
  on Empirical Methods in Natural Language Processing}}.
  \bibinfo{pages}{4772--4777}.
\newblock


\bibitem[\protect\citeauthoryear{Wang, Zhao, Liu, Chen, Zhang, and Wang}{Wang
  et~al\mbox{.}}{2021}]%
        {wang2021concept}
\bibfield{author}{\bibinfo{person}{Xuepeng Wang}, \bibinfo{person}{Li Zhao},
  \bibinfo{person}{Bing Liu}, \bibinfo{person}{Tao Chen}, \bibinfo{person}{Feng
  Zhang}, {and} \bibinfo{person}{Di Wang}.} \bibinfo{year}{2021}\natexlab{}.
\newblock \showarticletitle{Concept-Based Label Embedding via Dynamic Routing
  for Hierarchical Text Classification}. In
  \bibinfo{booktitle}{\emph{Proceedings of the 59th Annual Meeting of the
  Association for Computational Linguistics and the 11th International Joint
  Conference on Natural Language Processing}}. \bibinfo{pages}{5010--5019}.
\newblock


\bibitem[\protect\citeauthoryear{Xu, Teng, Zhao, Guo, Xiao, Jiang, and Ren}{Xu
  et~al\mbox{.}}{2021}]%
        {xu2021hierarchical}
\bibfield{author}{\bibinfo{person}{Linli Xu}, \bibinfo{person}{Sijie Teng},
  \bibinfo{person}{Ruoyu Zhao}, \bibinfo{person}{Junliang Guo},
  \bibinfo{person}{Chi Xiao}, \bibinfo{person}{Deqiang Jiang}, {and}
  \bibinfo{person}{Bo Ren}.} \bibinfo{year}{2021}\natexlab{}.
\newblock \showarticletitle{Hierarchical Multi-label Text Classification with
  Horizontal and Vertical Category Correlations}. In
  \bibinfo{booktitle}{\emph{Proceedings of the 2021 Conference on Empirical
  Methods in Natural Language Processing}}. \bibinfo{pages}{2459--2468}.
\newblock


\bibitem[\protect\citeauthoryear{Xu and Barbosa}{Xu and Barbosa}{2018}]%
        {xu2018neural}
\bibfield{author}{\bibinfo{person}{Peng Xu} {and} \bibinfo{person}{Denilson
  Barbosa}.} \bibinfo{year}{2018}\natexlab{}.
\newblock \showarticletitle{Neural Fine-Grained Entity Type Classification with
  Hierarchy-Aware Loss}. In \bibinfo{booktitle}{\emph{Proceedings of the 2018
  Conference of the North American Chapter of the Association for Computational
  Linguistics: Human Language Technologies}}. \bibinfo{pages}{16--25}.
\newblock


\bibitem[\protect\citeauthoryear{Yang, Sun, Li, Ma, Wu, and Wang}{Yang
  et~al\mbox{.}}{2018}]%
        {yang2018sgm}
\bibfield{author}{\bibinfo{person}{Pengcheng Yang}, \bibinfo{person}{Xu Sun},
  \bibinfo{person}{Wei Li}, \bibinfo{person}{Shuming Ma}, \bibinfo{person}{Wei
  Wu}, {and} \bibinfo{person}{Houfeng Wang}.} \bibinfo{year}{2018}\natexlab{}.
\newblock \showarticletitle{SGM: Sequence Generation Model for Multi-label
  Classification}. In \bibinfo{booktitle}{\emph{Proceedings of the 27th
  International Conference on Computational Linguistics}}.
  \bibinfo{pages}{3915--3926}.
\newblock


\bibitem[\protect\citeauthoryear{Zhou, Ma, Long, Xu, Ding, Zhang, Xie, and
  Liu}{Zhou et~al\mbox{.}}{2020}]%
        {zhou2020hierarchy}
\bibfield{author}{\bibinfo{person}{Jie Zhou}, \bibinfo{person}{Chunping Ma},
  \bibinfo{person}{Dingkun Long}, \bibinfo{person}{Guangwei Xu},
  \bibinfo{person}{Ning Ding}, \bibinfo{person}{Haoyu Zhang},
  \bibinfo{person}{Pengjun Xie}, {and} \bibinfo{person}{Gongshen Liu}.}
  \bibinfo{year}{2020}\natexlab{}.
\newblock \showarticletitle{Hierarchy-aware global model for hierarchical text
  classification}. In \bibinfo{booktitle}{\emph{Proceedings of the 58th Annual
  Meeting of the Association for Computational Linguistics}}.
  \bibinfo{pages}{1106--1117}.
\newblock


\end{thebibliography}

\end{document}